\documentclass[letterpaper, 10 pt, conference]{ieeeconf}
\IEEEoverridecommandlockouts
\usepackage{graphics} 
\usepackage{wrapfig} 
\usepackage[linesnumbered,ruled]{algorithm2e}
\graphicspath{{img/}}
\usepackage{epsfig} 
\usepackage{amsmath} 
\usepackage{amssymb}  
\usepackage{amsthm}
\usepackage{bigints}
\usepackage{pifont}
\def\-{\raisebox{.75pt}{-}}
\usepackage{mathtools}
\usepackage{cite}
\usepackage[makeroom]{cancel}

\usepackage[table]{xcolor} 

\usepackage[center]{subfigure}
\usepackage{multirow}
\usepackage{hyperref}
\usepackage{booktabs}
\usepackage{subfigure}
\usepackage{epstopdf}
\usepackage{textcomp}
\usepackage{makecell}
\usepackage{bm}
\usepackage{caption}
\usepackage{algpseudocode,algorithm} 
\usepackage{color}
\usepackage{mathptmx}
\usepackage[11pt]{moresize}
\usepackage{hhline}
\usepackage{tikz}
\usepackage{textcomp}
\def\BibTeX{{\rm B\kern-.05em{\sc i\kern-.025em b}\kern-.08em
    T\kern-.1667em\lower.7ex\hbox{E}\kern-.125emX}}

\newcommand{\subparagraph}{}
\usepackage{titlesec}
\titlespacing*{\section}
{0pt}{2mm plus0.7mm minus0.7mm}{2mm plus0.7mm minus0.7mm}
\titlespacing*{\subsection}
{0pt}{2mm plus0.7mm minus0.7mm}{2mm plus0.7mm minus0.7mm}
\titlespacing*{\subsubsection}
{0pt}{1mm plus0.7mm minus0.7mm}{1mm plus0.7mm minus0.7mm}

\title{\LARGE Can We Trust You? On Calibration of a Probabilistic Object Detector for Autonomous Driving
\thanks{$^1$ Robert Bosch GmbH, Corporate Research, Driver Assistance Systems and Automated Driving, 71272 Renningen, Germany.}
\thanks{$^2$ Institute of Measurement, Control and Microtechnology, Ulm University, 89081 Ulm, Germany.}
\thanks{We thank our colleagues Florian Faion and Florian Drews for their suggestions and inspiring discussions. We also thank Bill Beluch for reading the script. The video to this paper can be found at \url{https://youtu.be/pH5qT11vmyM}.}}

\author{Di Feng$^{1,2}$, Lars Rosenbaum$^1$, Claudius Gl\"aser$^1$, Fabian Timm$^1$, Klaus Dietmayer$^2$}

\let\oldtwocolumn\twocolumn
\renewcommand\twocolumn[1][]{%
	\oldtwocolumn[{#1}{
		\begin{center}
			\includegraphics[width=\textwidth]{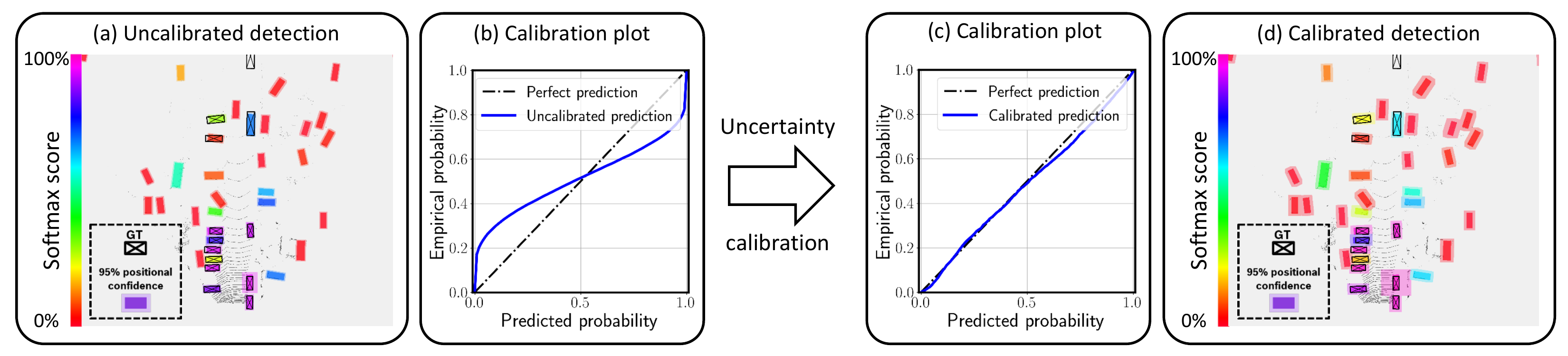}
			\captionof{figure}{A state-of-the-art probabilistic LiDAR 3D object detector produces uncalibrated uncertainties. \textbf{(a).} Each detection in the LiDAR Bird's eye view plane is colorized according to the softmax score. The $95\%$ positional confidence intervals in the horizontal plane (marginals) are drawn as shaded areas around each detection. \textbf{(b).} We visualize the uncertainty miscalibration problem by the calibration plot. \textbf{ (c).} Using our proposed uncertainty recalibration techniques, we significantly improve the uncertainty estimation quality. \textbf{(d).} The detection results after uncertainty recalibration.}\label{fig:framework}
			\vspace{+0.4em}
		\end{center}
	}]
}

\begin{document}
\maketitle

\begin{abstract}
Reliable uncertainty estimation is crucial for perception systems in safe autonomous driving. Recently, many methods have been proposed to model uncertainties in deep learning-based object detectors. However, the estimated probabilities are often uncalibrated, which may lead to severe problems in safety-critical scenarios. In this work, we identify such uncertainty miscalibration problems in a probabilistic LiDAR 3D object detection network, and propose three practical methods to significantly reduce errors in uncertainty calibration. Extensive experiments on several datasets show that our methods produce well-calibrated uncertainties, and generalize well between different datasets.
\end{abstract}


\section{Introduction}\label{sec:introduction}
Reliable uncertainty estimation in object detection systems is crucial for safe autonomous driving. Intuitively, a probabilistic object detector should predict uncertainties that match the natural frequency of correct predictions. For example, if the detector makes predictions with $0.9$ probability, then $90\%$ of those predictions should be correct. Reliable uncertainty estimation builds trust between a driverless car and its users, as humans have an intuitive notion of probabilities in a frequentist sense~\cite{cosmides1996humans}. Moreover, the uncertainties captured in object detectors can be propagated to other modules, such as tracking and motion planning~\cite{banzhaf2018footprints}, so that the overall system robustness can be enhanced. 

In recent years, many methods have been proposed to model uncertainties in deep neural networks. Among them, the direct-modeling approach assumes a certain probability distribution over the network outputs (e.g. Gaussian distribution), and uses additional output layers to predict parameters for such a distribution. Due to its simplicity and real-time implementation, this method has been widely applied to object detectors in autonomous driving~\cite{feng2018towards,feng2018leveraging,harakeh2019bayesod,le2018uncertainty,wirges2019capturing,meyer2019lasernet}. However, we find that the direct-modeling approach fails to produce reliable probabilities, causing uncertainty miscalibration problems. Using such an unreliable uncertainty estimation in object detectors can lead to wrong decision makings in autonomous driving (e.g. at the planning stage), which may cause fatal accidents, especially in safety-critical scenarios. 

In this study, we identify uncertainty miscalibration problems in a probabilistic LiDAR object detection network (Sec.~\ref{sec:object_detector}) via calibration plots (Sec.~\ref{sec:uncertainty_evaluation}). Then, we propose three practical methods based on recalibration techniques to alleviate such miscalibration (Sec.~\ref{sec:uncertainty_recalibration}), and systematically study their robustness on several datasets. Experimental results show that our methods can significantly reduce the uncertainty calibration errors and improve the detection accuracy (Sec.~\ref{sec:result}). 

\section{Related Work}
\label{sec:related_works}
\subsection{Uncertainty Estimation for Object Detection}
The methods to model uncertainty in object detection can be categorized into two groups: the ensemble approach and the direct-modeling approach. The former builds an ensemble of object detectors to \textit{approximate} an output probability distribution with samples, e.g. using Monte-Carlo Dropout~\cite{Gal2016Uncertainty}. This approach has shown to represent the model uncertainty, and has successfully been introduced to tackle open-set object detection challenges~\cite{miller2017dropout,miller2018evaluating} and active learning~\cite{feng2019deep2}. The direct-modeling approach uses network output layers to learn and predict the parameters of a \textit{pre-defined} probability distribution, such as a multi-variate Gaussian distribution~\cite{feng2018towards,feng2018leveraging} or mixture of Gaussian~\cite{meyer2019lasernet}. It requires only a little additional computation during inference, and can improve the detection accuracy~\cite{feng2018leveraging}. Therefore, we employ the direct-modeling approach to model uncertainty in our LiDAR object detector.

\subsection{Uncertainty Recalibration}
Uncertainty recalibration techniques aim to improve the uncertainty estimation of a probabilistic model. Most of them are post-processing steps that directly adjust network probabilistic outputs via a recalibration model. Many models have been developed to calibrate classification uncertainty in deep learning, such as isotonic regression, histogram binning, and temperature scaling~\cite{guo2017calibration}. Besides, \cite{kuleshov2018accurate} introduces isotonic regression to calibrate uncertainties in multiple regression tasks. To the best of our knowledge, there is no previous work focusing on how to calibrate uncertainties for object detections. In this work, we employ isotonic regression and temperature scaling to recalibrate the classification part of our object detector, and extend them to recalibrate uncertainties over the bounding box predictions. Furthermore, we propose a simple loss function to reduce calibration errors during training.
\section{Probabilistic LiDAR Object Detection}
\label{sec:object_detector}
\subsection{Network Architecture}\label{subsec:network_architecture}
\begin{figure}[H]
	\centering
	\begin{minipage}{0.8\linewidth}
		\centering
		\includegraphics[width=1\linewidth]{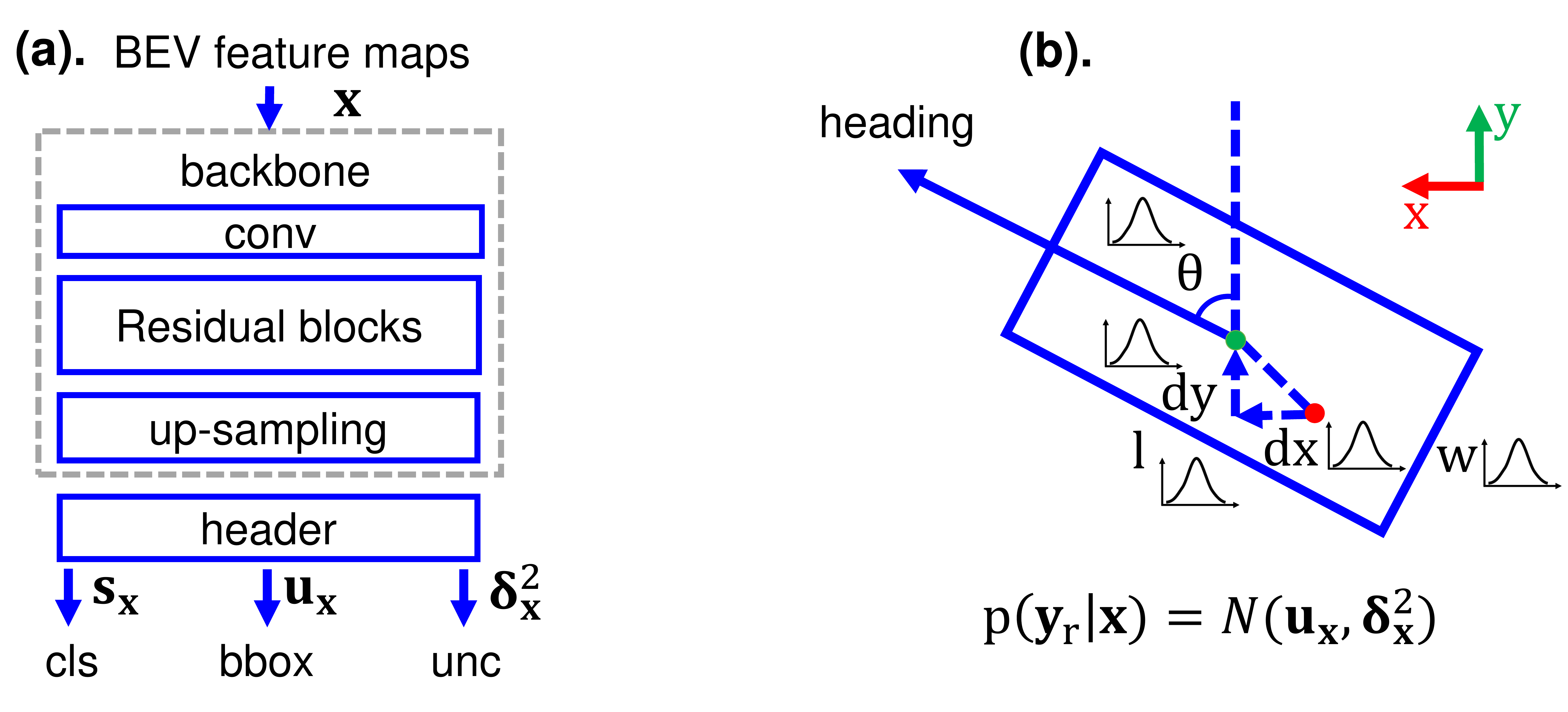}
	\end{minipage}
	\caption{\textbf{(a).} Network architecture; \textbf{(b).} Bounding box encoding with probability estimation.}\label{fig:architecture}
	\vspace{-1em}
\end{figure}

We model uncertainties in PIXOR~\cite{Yang_2018_CVPR}, a state-of-the-art one-stage LiDAR object detection network, with several modifications (Fig.~\ref{fig:architecture})~\cite{feng2018leveraging}. PIXOR takes the LiDAR bird's eye view (BEV) feature maps as input, and outputs classification scores and bounding box parameters for each pixel on the feature map. Denote an input sample as $\mathbf{x}$, the network predicts object classes $y_c$ with softmax score $
s_{\mathbf{x}}$ (for brevity we only consider binary classification ``Object" and ``Background", i.e. $y_c \in\{0,1\}$). It also regresses the object positions $\mathbf{y}_r \in \mathbb{R}^6$ including center positional offsets on the horizontal plane ($dx$ and $dy$), length $l$, width $w$, and orientation $\theta$. Following~\cite{Yang_2018_CVPR}, we encode the the final bounding box locations as the \textit{row vector} $\mathbf{u}_{\mathbf{x}} = [\cos(\theta), \sin(\theta), dx, dy, \log(l), \log(w)]_{\mathbf{x}}$ (Fig.~\ref{fig:architecture}).

\subsection{Uncertainty Estimation}\label{subsec:uncertainty_estimation}
We leverage our previously proposed method~\cite{feng2018leveraging} to model uncertainties in the object detection network. The softmax score is used to estimate the object probability, i.e. $p(y_c=1|\mathbf{x})=s_{\mathbf{x}}$. We also assume that the network regression outputs follow a multi-variate Gaussian distribution $p(\mathbf{y}_r|\mathbf{x}) = \mathcal{N}(\mathbf{u}_{\mathbf{x}}, \Sigma_{\mathbf{x}})$, with its mean being the network's bounding box regression $\mathbf{u}_{\mathbf{x}} = [\cos(\theta), \sin(\theta), dx, dy, \log(l), \log(w)]_{\mathbf{x}}$, and its covariance matrix being a diagonal matrix: $\Sigma_{\mathbf{x}} = \mathrm{diag}(\mathbf{\sigma}^2_{\mathbf{x}})$, $\mathbf{\sigma}^2_{\mathbf{x}} = [\sigma^2_{\cos(\theta)}, \sigma^2_{\sin(\theta)}, \sigma^2_{dx}, \sigma^2_{dy}, \sigma^2_{\log(l)}, \sigma^2_{\log(w)}]_{\mathbf{x}}$. Here, each element in the \textit{row vector} $\mathbf{\sigma}^2_{\mathbf{x}}$ represents a variance (or uncertainty) 
to the corresponding element in $\mathbf{u}_{\mathbf{x}}$. We add additional output layers in our object detector to directly regress $\mathbf{\sigma}^2_{\mathbf{x}}$. In this way, the network can estimate the probability distribution of the bounding box prediction during test time. We employ the multi-loss function~\cite{feng2018leveraging} to train the regression tasks:
\begin{equation}\label{eq:loss_attenuation}
L_{reg} = \frac{1}{2}(\mathbf{y}_r-\mathbf{u}_{\mathbf{x}})\mathrm{diag}(\frac{1}{\mathbf{\sigma}^2_{\mathbf{x}}})(\mathbf{y}_r-\mathbf{u}_{\mathbf{x}})^T + \frac{1}{2}\log(\mathbf{\sigma}^2_{\mathbf{x}})\mathbf{1}^T,
\end{equation}
with $\mathbf{y}_r$ being the regression ground truth. 


\section{Uncertainty Evaluation}
\label{sec:uncertainty_evaluation}
\subsection{Definition}
Let a labeled dataset $\{(\mathbf{x}^n, y_c^n, \mathbf{y}_r^n)\}_{n=1}^N$ be the i.i.d. realizations of jointly distributed random variables $X$, $Y_c$ and $Y_r$, where $X$ refers to the input data, $Y_c$ the binary classification labels $\{0,1\}$, and $Y_r$ the bounding box locations. The marginal distributions of the target variables can be specified as $X, Y_c \sim \mathbb{P}_c$ and $X, Y_r \sim \mathbb{P}_r$. Our LiDAR detector predicts the softmax score as the object probability distribution, which can be denoted as $F^n_c(y_c=1)=p(y_c=1|\mathbf{x}^n)=s_{\mathbf{x}^n}$. In the case of bounding box regression, we use the cumulative distribution function (CDF) $F^n_r(\mathbf{y}_r)$, with its probability density function being predicted by the detector $p(\mathbf{y}^n_r|\mathbf{x}^n) = \mathcal{N}(\mathbf{u}_{\mathbf{x}^n}, \Sigma_{\mathbf{x}^n})$, and $F^{n^{-1}}_r(p)$ for its quantile function (inverse cumulative distribution function).

Intuitively, a reliable (or calibrated) uncertainty estimation from the object detector means that a predicted probability should match the natural frequency of correct predictions. For example, if the detector classifies $100$ objects as ``Car" with a softmax score of $0.9$, we expect that $90\%$ of those objects should be correctly classified; In the regression task, if the detector estimates $100$ object positions with the $90\%$ confidence interval, $90\%$ of object ground truth positions should fall into such a confidence interval. More formally, a calibrated classification uncertainty for the class label $k$ indicates $\mathbb{P}_c\big(Y_c=1|F_c(Y_c=1)=p\big) = p,\ \ \forall p \in [0,1]$.
When $N$ is large enough, we have the following approximation~\cite{guo2017calibration}:
\begin{equation}\label{eq:classification_prob}
\mathbb{P}_c\big(Y_c=1|F_c(Y_c=1)=p\big) \approx \frac{\sum_{n=1}^N \mathbb{I}\big(y^n_c=1, F^n_c(y_c=1)=p\big)}{\sum_{n=1}^N \mathbb{I}\big(F^n_c(y_c=1)=p\big)}.
\vspace{-0.5em}
\end{equation}

\vspace{1em}
Similarly, a calibrated regression uncertainty means $\mathbb{P}_r\big(Y_r \leq F^{-1}_r(p)) = p,\ \ \forall p \in [0,1]$. It can be approximated by~\cite{kuleshov2018accurate}:
\begin{equation}\label{eq:regression_prob}
\mathbb{P}_r\big(Y_r \leq F^{-1}_r(p)) \approx \frac{\sum_{n=1}^N \mathbb{I}\big(\mathbf{y}^n_r \leq F^{n^{-1}}_r(p) \big)}{N}.
\end{equation}

By using a multi-variate Gaussian distribution with the diagonal covariance matrix, we assume that each element in the regression task is independently distributed. In this regard, we only focus to calibrate the \textit{marginal} probability for each element separately. In the rest of the paper, we denote $y_r$ as an element in the regression task, and $F^n_r(y_r)$ its CDF. We leave a comprehensive study to calibrate the full probability, where the elements in the regression are dependent, as an interesting future work.
\subsection{Evaluation Tools}
The notion of uncertainty calibration can be visualized by the calibration plot (Fig.~\ref{fig:framework}), where the horizontal axis represents the predicted probability of the detector, and the vertical axis the empirical probability. In practice, to draw calibration plot for classification, we group the softmax scores into $M$ intervals using the probability thresholds $0 < p_c^1 < ... < p_c^m < ... < 1$, and calculate the empirical probability following Eq.~\ref{eq:classification_prob} for each interval, denoted as $\hat{p}_c^m$. In the case of regression, we group predictions into different confidence levels $p_r^m$, calculated by $F_r(\mathbf{y}_r)$, and estimate the corresponding empirical frequency $\hat{p}_r^m$ by Eq.~\ref{eq:regression_prob}. Here, we draw a calibration plot for each bounding box regressor separately. A well-calibrated detector produces the diagonal line in the calibration plot (Fig.~\ref{fig:framework}(b)). A miscalibrated detector can suffer from over-confident predictions (actual calibration curve is under the diagonal line), or under-confident predictions (above the diagonal line).

Similar to~\cite{guo2017calibration}, we employ Expected Calibration Error (ECE) as evaluation matric. ECE calculates the weighted error between the actual calibration curve and the diagonal line, i.e. $\text{ECE}=\sum_{m=1}^M \frac{N_m}{N}|p^m-\hat{p}^m|$, with $N_m$ being the number of samples in the $m$th interval. $\text{ECE}=0$ corresponds to perfectly calibrated predictions.

\section{Uncertainty Recalibration}
\label{sec:uncertainty_recalibration}
In this section, we introduce in detail our methods to separately recalibrate the marginal probability distribution in each element of the bounding box regression.

\subsection{Isotonic Regression}
Recall that $p = F_r(y_r)$ represents the predicted bounding box probability from the network. We train an auxiliary model based on the isotonic regression $p \mapsto g(p)$, which is a non-parametric monotonically increasing function, to fit the true probability $\mathbb{P}_r\big(Y_r \leq F^{-1}_r(p))$~\cite{kuleshov2018accurate}. During test time, the object detector produces an uncalibrated uncertainty, which will then be corrected by the recalibration model $g(\cdot)$ as the final output. In practice, we build a recalibration dataset from the validation data to learn the isotonic regression model for our pre-trained LiDAR object detector. Specifically, denote $\{(\mathbf{x}^n, \mathbf{y}_r^n)\}_{n=1}^N$ as the validation dataset; we can build its corresponding recalibration data $\{\big( F^n_r(\mathbf{y}^n_r), \hat{P}(F_r(\mathbf{y}_r)) \big) \}_{n=1}^N$, where $\hat{P}\big(F_r(\mathbf{y}_r)\big)$ refers to the empirical probability calculated by Eq.~\ref{eq:regression_prob}. 

\subsection{Temperature Scaling}
We use different scalars $T>0$ for \textit{each} regressor to adjust the variance prediction:
$\hat{\sigma} \leftarrow \sigma^2/T, \ \ \forall \sigma^2 \in \{\sigma^2_{\cos(\theta)}, \sigma^2_{\sin(\theta)}, \sigma^2_{dx}, \sigma^2_{dy}, \sigma^2_{\log(l)}, \sigma^2_{\log(w)}\}$. When $T>1$, the adjusted Gaussian distribution becomes sharper, indicating smaller uncertainty. When $T<1$, the distribution becomes broader, representing larger uncertainty. With $T=1$, the original probability is unchanged. The optimal $T$ can be found by maximizing the Negative Log Likelihood (NLL) score on the recalibration dataset. 

\subsection{Calibration Loss} 
The object detector learns to predict variances in an unsupervised way (Eq.~\ref{eq:loss_attenuation}), as there is no ground truth for variances in Eq.~\ref{eq:loss_attenuation}. Therefore, the regression loss function is not designed to guarantee calibrated uncertainty. Intuitively, a well-calibrated uncertainty for a Gaussian distribution indicates that for every data sample $\mathbf{x}$, the predicted variances should match the true differences between predicted and true bounding boxes, i.e. $\sigma^2_{\mathbf{x}} \overset!= (\mathbf{y}_r - \mathbf{u}_{\mathbf{x}}) \odot (\mathbf{y}_r - \mathbf{u}_{\mathbf{x}})$, with $\odot$ being the element-wise multiplication. In this regard, we design a simple calibration loss $L_{calib}$ to regularize variances, and train the object detector with a new loss function $L_{total}$ that adds $L_{reg}$ and $L_{calib}$:
\begin{equation}\label{eq:calibration_loss}
\begin{split}
& L_{calib} = \|\sigma^2_{\mathbf{x}} - (\mathbf{y}_r - \mathbf{u}_{\mathbf{x}}) \odot (\mathbf{y}_r - \mathbf{u}_{\mathbf{x}}) \|, \\
& L_{total} = L_{reg} + \lambda L_{calib},
\end{split}
\end{equation}
where the hyper-parameter $\lambda$ is used to control the loss weight.
    
\subsection{Comparison}\label{subsec:comparison}
All three uncertainty recalibration methods can improve the probability estimations (Sec.~\ref{subsubsec:uncertainty_recalibration:performance}). Isotonic regression and temperature scaling are post-processing steps after training the object detector. They do not change mean values $\mathbf{u}_{\mathbf{x}}$, and thus do not affect the detection accuracy. Furthermore, they are designed to optimize the uncertainty estimation based on the whole recalibration dataset, and do not guarantee that each detection is better-calibrated. Conversely, calibration loss improves the probability estimations when optimizing the object detector. It indirectly improves the detection accuracy by encouraging the network to produce better-calibrated uncertainties for \textit{each} detection (Sec.~\ref{subsubsec:uncertainty_recalibration:performance}). 

Given enough recalibration data, isotonic regression is guaranteed to produce perfect calibration plots, regardless of the underlying probability distributions~\cite{kuleshov2018accurate}. However, it changes the probability distribution (in our case a non-Gaussian distribution), making it less interpretable and applicable. On the contrary, temperature scaling and calibration loss recalibrate uncertainties based on the same probability distribution (in our case a Gaussian distribution), which is highly desirable when propagating them to other modules, such as object tracking with Kalman filters. However, if the assumed probability distribution significantly differs from the true distribution, both recalibration methods may fail to achieve well-calibrated uncertainties.

\section{Experimental Results}
\label{sec:result}

\begin{table*}
	\centering
	\makebox[0pt][c]{\parbox{1.00\textwidth}{%
			\begin{minipage}[bt]{0.35\hsize}\centering
				\resizebox{0.95\linewidth}{!}{
					\begin{tabular}{ c | c c c}
					 	\rowcolor{lightgray!75} \textbf{\texttt{Network}} & \textbf{\texttt{Easy}} & \textbf{\texttt{Moderate}} & \textbf{\texttt{Hard}}  \\
						\textbf{\texttt{PIXOR~\cite{Yang_2018_CVPR}}} & $86.79$ & $80.75$ & $76.60$ \\
						\textbf{\texttt{Ours}}  & $87.48$ & $78.29$ & $75.41$ \\
						\textbf{\texttt{Ours + Calib. Loss}} & $\mathbf{90.91}$ & $\mathbf{81.81}$ & $\mathbf{79.12}$ \\
				\end{tabular}}
				\caption{Detection performance (Average Precision on the Bird's Eye View: $AP_{BEV} (\%)$) on the KITTI \textit{val} set.}
				\label{tab:detection_performance}
				\vspace{-2em}
			\end{minipage}
			\hfill
			\begin{minipage}[bt]{0.65\hsize}\centering
				\resizebox{0.95\linewidth}{!}{
					\begin{tabular}{c | l l l l l l l | l }
						\rowcolor{lightgray!75} \textbf{\texttt{Method}} & cls & $\cos(\theta)$ & $\sin(\theta)$ & $dx$ & $dy$ & $\log(w)$ & $\log(l)$ & \textbf{\texttt{avg.}} \\ 
						
						\textbf{\texttt{Uncalibrated (Baseline)}} & $0.109 $ & $0.092$ & $0.117$ & $0.141$ & $0.179$ & $0.286$ & $0.186$ & $0.159$ \\
						
						\textbf{\texttt{Calib. Loss}} & $0.101$ & $0.067$ & $0.091$ & $0.131$ & $0.121$ & $0.142$ & $0.152$ & $0.115$ \\ 
						
						\textbf{\texttt{Temp. Scaling}} & $0.041$ & $0.079$ & $0.029$ & $0.025$ & $0.037$ & $0.126$ & $0.078$ & $0.059$ \\
						
						\textbf{\texttt{Isotonic Regr.}} & $0.005$ & $0.014$ & $\mathbf{0.003}$ & $0.015$ & $0.018$ & $0.007$ & $0.018$ & $0.011$ \\ 
						
						\textbf{\texttt{Calib. Loss + Temp. Scaling}} & $0.048$ & $0.067$ & $0.016$ & $0.021$ & $0.018$ & $0.018$ & $0.060$ & $0.035$ \\
						
						\textbf{\texttt{Calib. Loss + Isotonic Regr.}} & $\mathbf{0.004}$ & $\mathbf{0.005}$ & $0.004$ & $\mathbf{0.007}$ & $\mathbf{0.007}$ & $\mathbf{0.004}$ & $\mathbf{0.003}$ & $\mathbf{0.005}$\\ 
					\end{tabular}}
				\caption{Expected calibration errors (ECE) on the KITTI \textit{eval} set.}\label{tab:recalibration_performance}
				\vspace{-2em}
			\end{minipage}
		}}
\end{table*}

\subsection{Identifying Uncertainty Miscalibration} \label{subsec:uncertainty_miscalibration}
In the first experiment, we use calibration plots to identify the uncertainty miscalibration problem in our probabilistic LiDAR object detector, and then study how such a problem is related to the training process. For the sake of brevity, here we only show the marginal of the $dy$ regression variable, though we observe similar results in other regression variables as well. 

\subsubsection{Experimental Setup}
We conduct experiments on the training data of the KITTI object detection benchmark~\cite{geiger2012we} using only the ``Car" category. We split the data into a \textit{train} set and a \textit{val} set with approximately 50:50 ratio~\cite{chen2016multi}. The LiDAR detector is trained with the KITTI \textit{train} set, and its uncertainty estimation quality is evaluated on the KITTI \textit{val} set. Similar to~\cite{feng2018leveraging}, we pre-train the detector with the normal $L_2$ loss for $45$ epochs, using the SGD optimizer with a learning rate of $0.02$. Then, we reduce the learning rate to $0.001$ and train the detector following Eq.~\ref{eq:loss_attenuation} for another $100$ epochs. Tab.\ref{tab:detection_performance} reports the car detection performance in the Bird's Eye View (BEV), with the Intersection Over Union IOU=$0.7$ threshold. Our network (``Ours") produces on-par results with the original PIXOR network.

\subsubsection{Calibration Plots}
\begin{figure}[htpb]
	\centering
	\begin{minipage}{0.98\linewidth}
		\centering
		\subfigure[Classification]{\label{fig:calibration_classification}\includegraphics[width=0.46\linewidth]{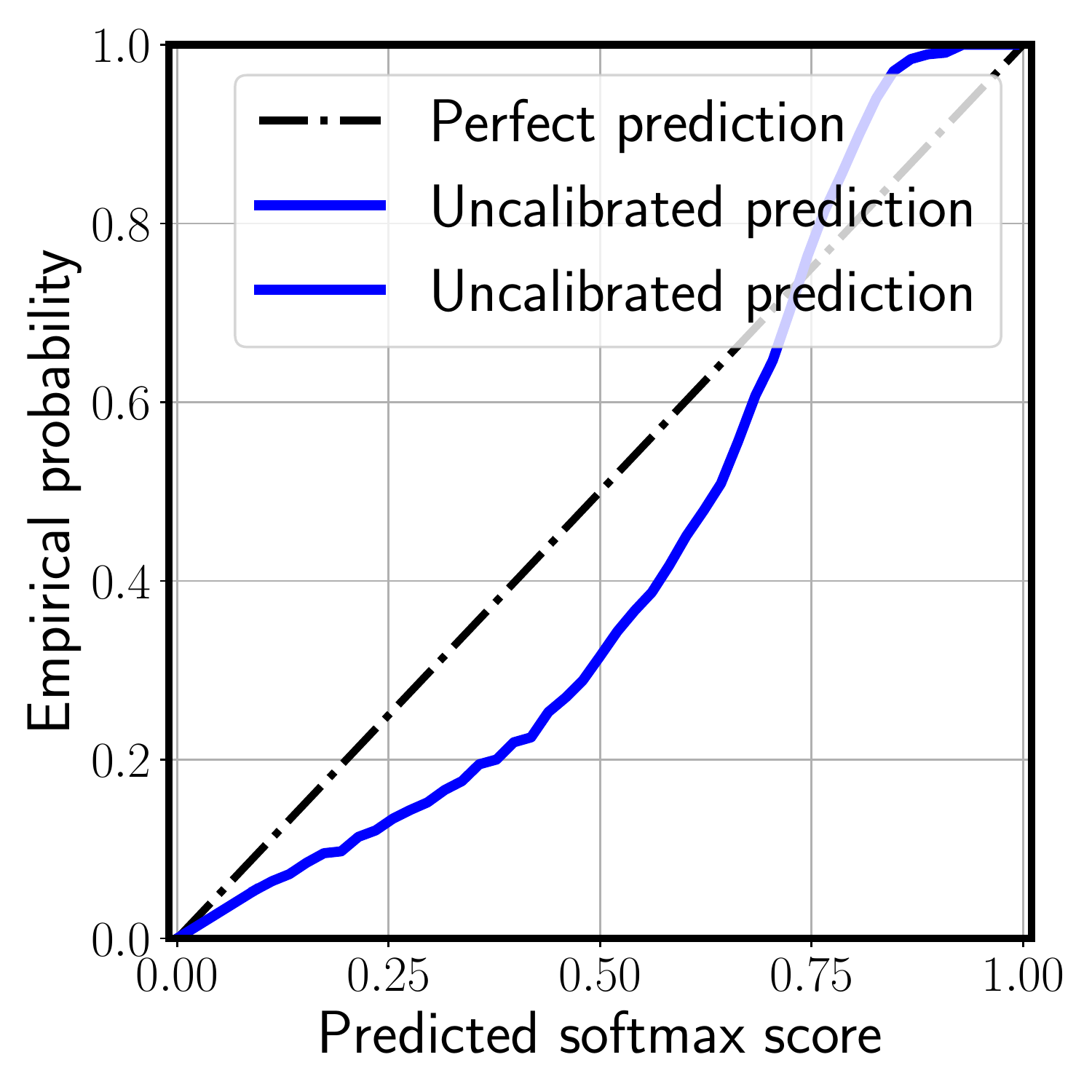}}
		\subfigure[Regression]{\label{fig:calibration_regression}\includegraphics[width=0.46\linewidth]{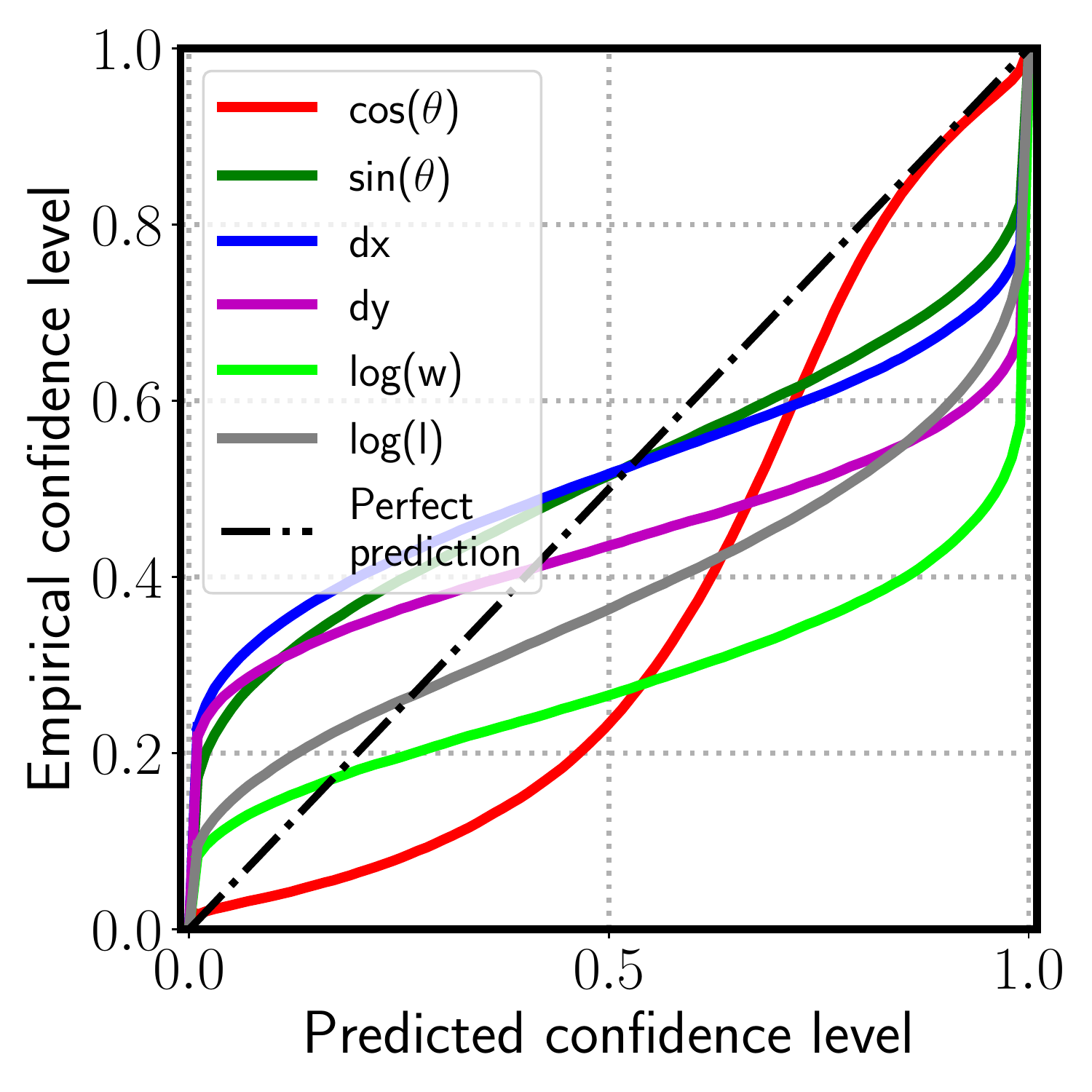}}
		\caption{Identifying the uncertainty miscalibration problem. \textbf{(a)} Calibration plot for the classification task. \textbf{(b)} Calibration plot for the marginal probability distributions for regression variables.}\label{fig:calibration_plot}
	\end{minipage}
	\vspace{-1em}
\end{figure}

Fig.~\ref{fig:calibration_plot} shows the calibration plots for classification and regression respectively. From the figures we observe that the probabilistic object detector produces miscalibrated uncertainties. For example, the network is over-confident classifying objects when the predicted softmax scores are smaller than $0.7$, and under-confident with softmax scores larger than $0.7$ (Fig.~\ref{fig:calibration_classification}). Conversely, except for the $\cos(\theta)$ regression, the network makes under-confident predictions at a smaller confidence levels, and over-confident predictions at a higher confidence levels (Fig.~\ref{fig:calibration_regression}). 

\subsubsection{Training Epochs}
\begin{figure}[tpb]
	\centering
	\begin{minipage}{0.6\linewidth}
		\centering
		\includegraphics[width=0.9\linewidth]{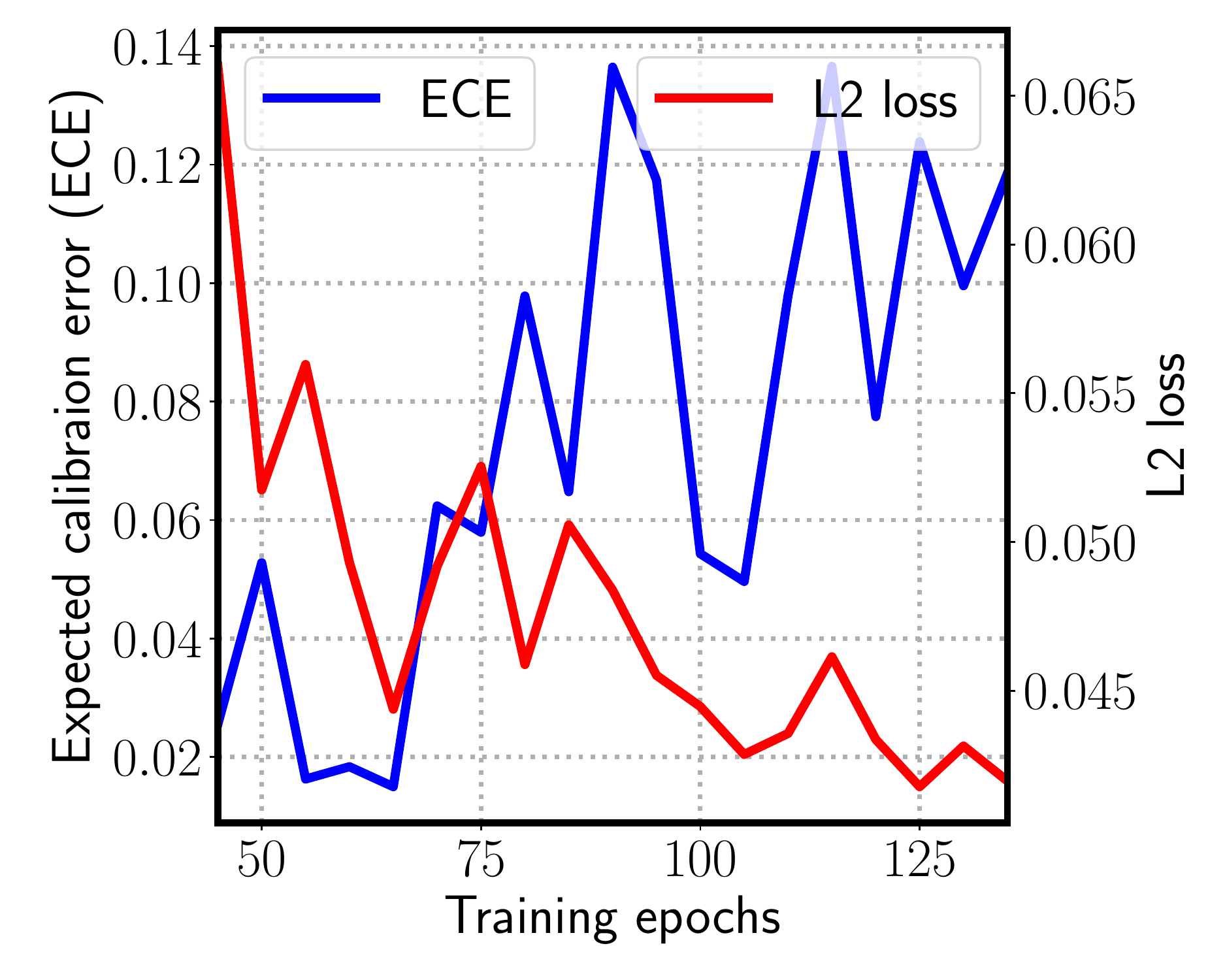}
	\end{minipage}
	\hskip 3pt
	\begin{minipage}{0.36\linewidth}
		\centering
		\caption{The evolution of calibration errors and $L_2$ loss ($L_{reg}$) fpr $dy$ prediction wrt. training steps. The horizontal axis begins at $45$ because we start to do probabilistic modeling at the $45$th training epoch. }\label{fig:training_epoch}
	\end{minipage}
	\vspace{-2em}
\end{figure}

We find that achieving more accurate object detection does not guarantee better uncertainty estimation. Fig.~\ref{fig:training_epoch} illustrates how the regression and uncertainty estimation errors for $dy$ predictions over the course of training. The horizontal axis represents the training epochs, starting at epoch 45, when we start to model regression uncertainties using Eq.~\ref{eq:loss_attenuation}. The vertical axis represents the expected calibration errors and $L_2$ loss calculated on the \textit{val} set. The $L_2$ loss drops during the training, indicating that the network makes more and more accurate $dy$ predictions. However, the calibration errors tend to increase after the $65$th training epoch, showing over-fitting behaviour. A similar phenomenon for classification is found by Guo \textit{et al.}~\cite{guo2017calibration}. 

\begin{figure*}[htpb]
	\centering
	\begin{minipage}{0.98\textwidth}
		\centering
		\subfigure[Uncalibrated predictions]{\label{fig:1}\includegraphics[width=0.23\textwidth]{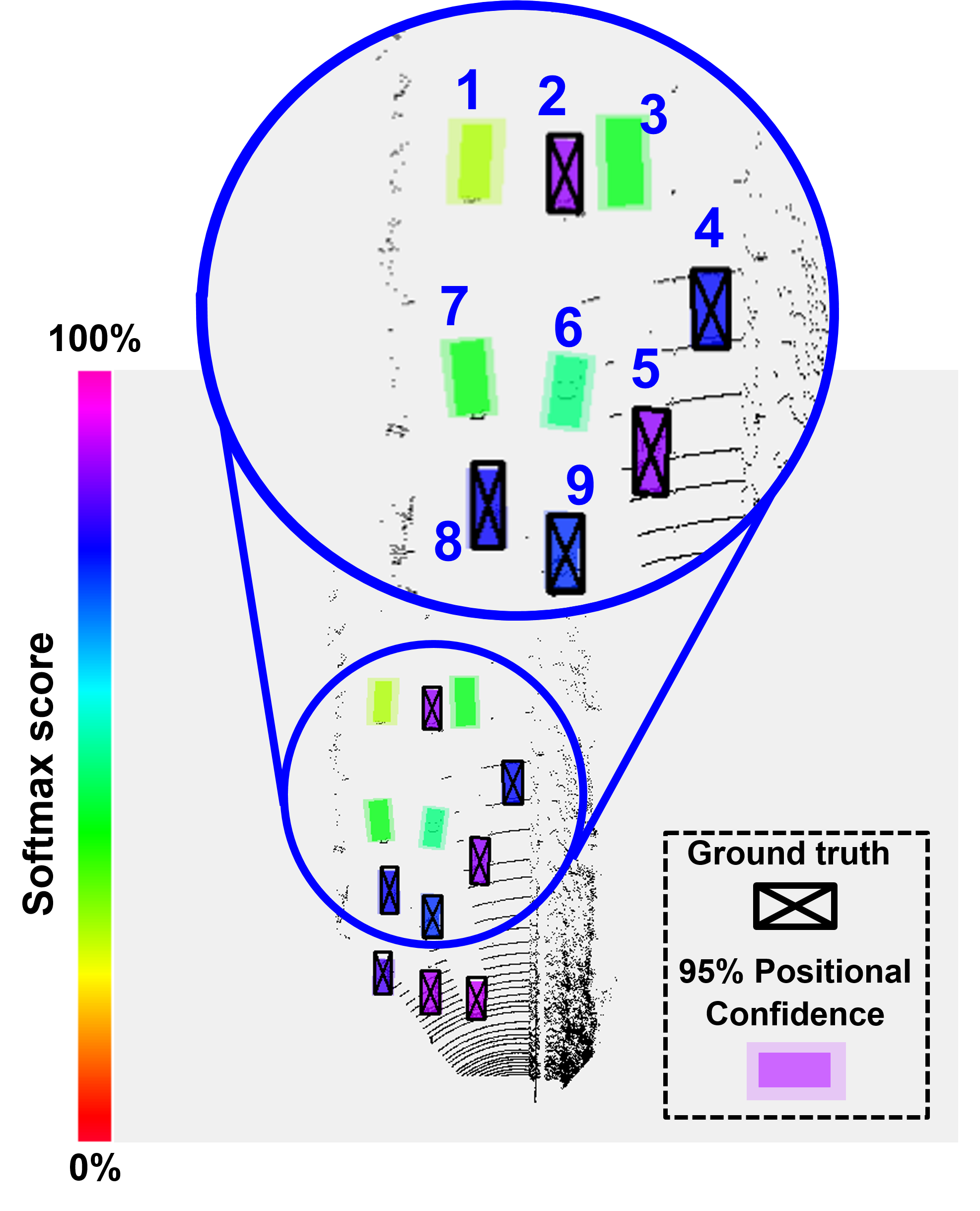}}
		\subfigure[Calibrated predictions]{\label{fig:2}\includegraphics[width=0.23\textwidth]{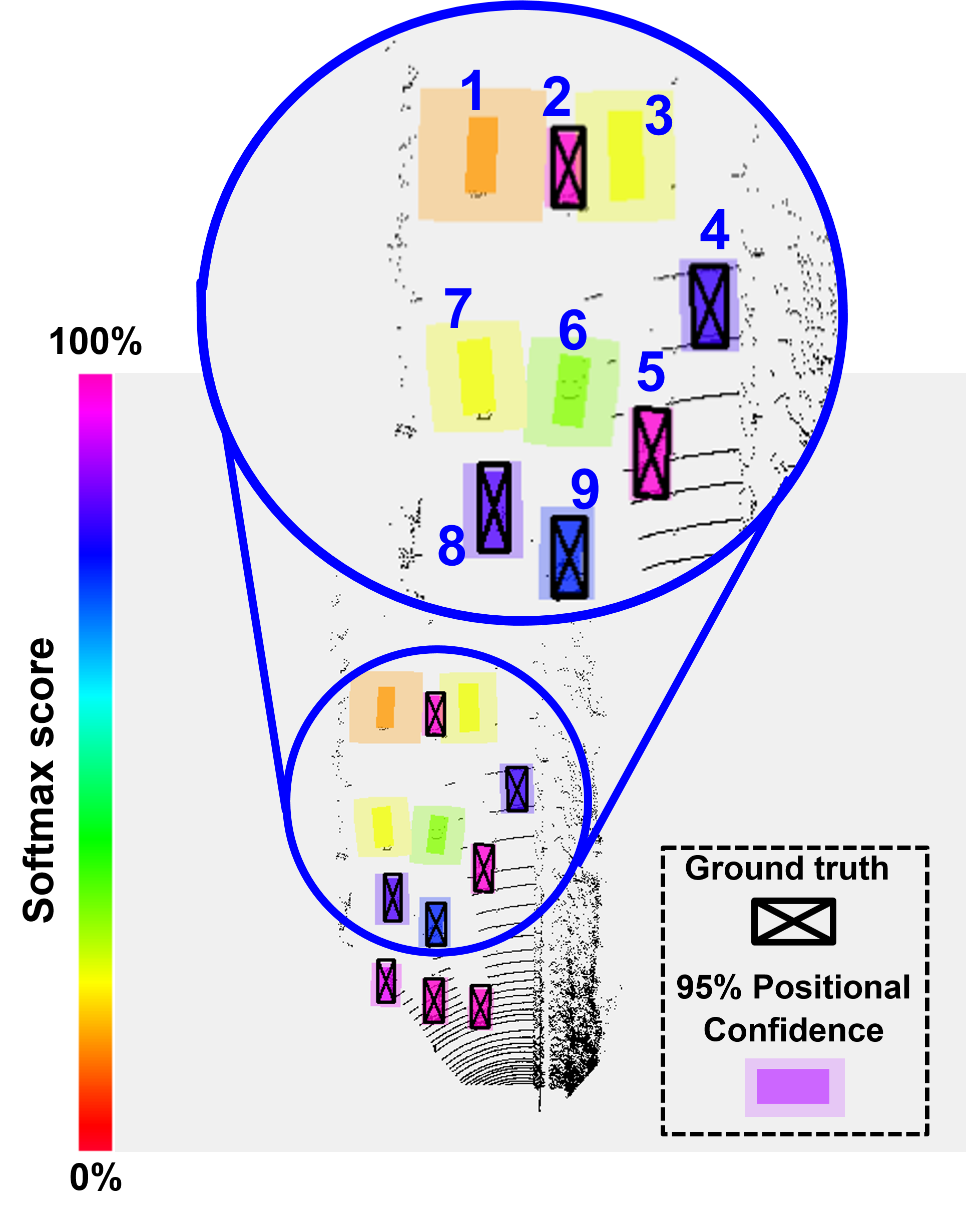}}
		\hfill
		\subfigure[Uncalibrated predictions]{\label{fig:3}\includegraphics[width=0.23\textwidth]{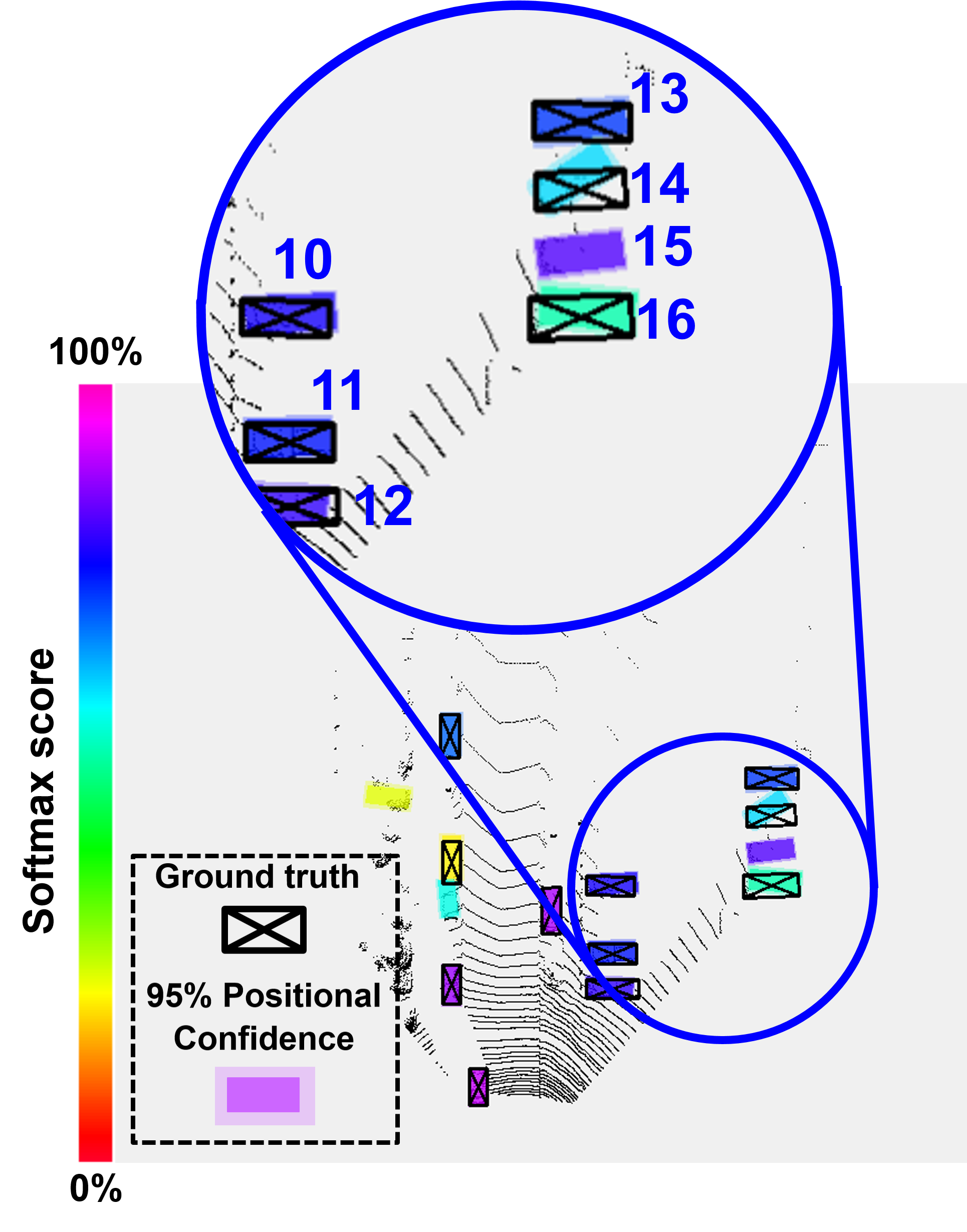}}
		\subfigure[Calibrated predictions]{\label{fig:4}\includegraphics[width=0.23\textwidth]{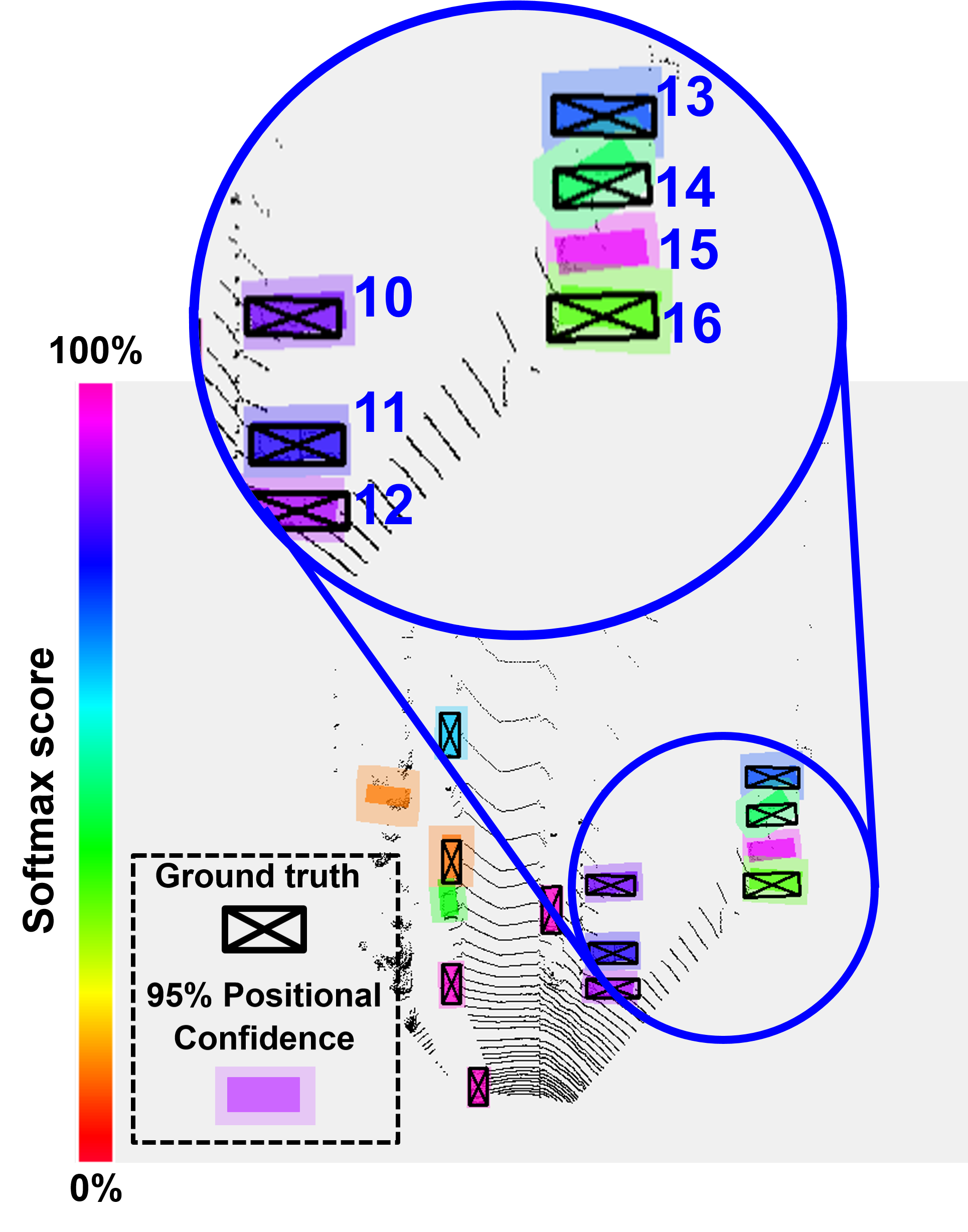}}
		\caption{Predictions with uncalibrated and recalibrated uncertainties.}\label{fig:qualitative_observation}
	\end{minipage}
	\vspace{-1em}
\end{figure*}

\subsection{Uncertainty Recalibration} \label{subsec:uncertainty_recalibration}
In this experiment, we evaluate the performance of the proposed uncertainty recalibration methods (``Temp. Scaling", ``Isotonic Regr.", and ``Calib. Loss"). We first show that the network produces better uncertainties after recalibration. It also achieves a higher detection accuracy when being trained with the calibration loss. Afterwards, we systematically study the robustness of the recalibration methods against different recalibration dataset sizes, and their generalization between different datasets.

\subsubsection{Experimental Setup}
The uncertainties predicted by the LiDAR detector in the previous experiment are used as the baseline. They are also used as inputs for the ``Temp. Scaling" and ``Isotonic Regr". For ``Calib. Loss", we train a new LiDAR detector with the calibration loss (Eq.~\ref{eq:calibration_loss}) using the KITTI \textit{train} set. Additionally, we combine the calibration loss with the other two recalibration methods as a comparison (``Calib. Loss + Temp. Scaling" and ``Calib. Loss + Isotonic Regr."). We equally split the KITTI \textit{val} set (Sec.~\ref{subsec:uncertainty_miscalibration}) into the KITTI \textit{recal} set and the KITTI \textit{eval} set. The former is used to optimize the recalibration models built by the temperature scaling and isotonic regression, and the latter to evaluate all recalibration methods. 

\subsubsection{Performance}\label{subsubsec:uncertainty_recalibration:performance}
Tab.~\ref{tab:detection_performance} shows that the network trained with the calibration loss (``Ours + Calib. Loss") improves the average precision (AP) up to nearly $4\%$ compared to the network without calibration loss (``Ours"). This might because calibration loss serves to regularize $\sigma^2$. As a result, the network learns to detect objects more accurately with improved uncertainty estimation. Tab.~\ref{tab:recalibration_performance} compares the Expected Calibration Errors (ECE) between the recalibrated uncertainties and the baseline uncertainties without recalibration. All recalibration methods consistently outperform the baseline with smaller ECE values. Specifically, ``Isotonic Regr." performs better than ``Temp. Scaling" and ``Calib. Loss". This is because the recalibration dataset is large enough to train a well-performed isotonic regression model (cf. Sec.~\ref{subsec:comparison} for more discussion). When combing ``Isotonic Regr." and ``Calib. Loss", we achieve the best calibrated uncertainties.  

\begin{figure}[htpb]
	\centering
	\begin{minipage}{0.98\linewidth}
		\centering
		\subfigure[Temp. Scaling]{\label{fig:num_data_temp_scaling}\includegraphics[width=0.49\textwidth]{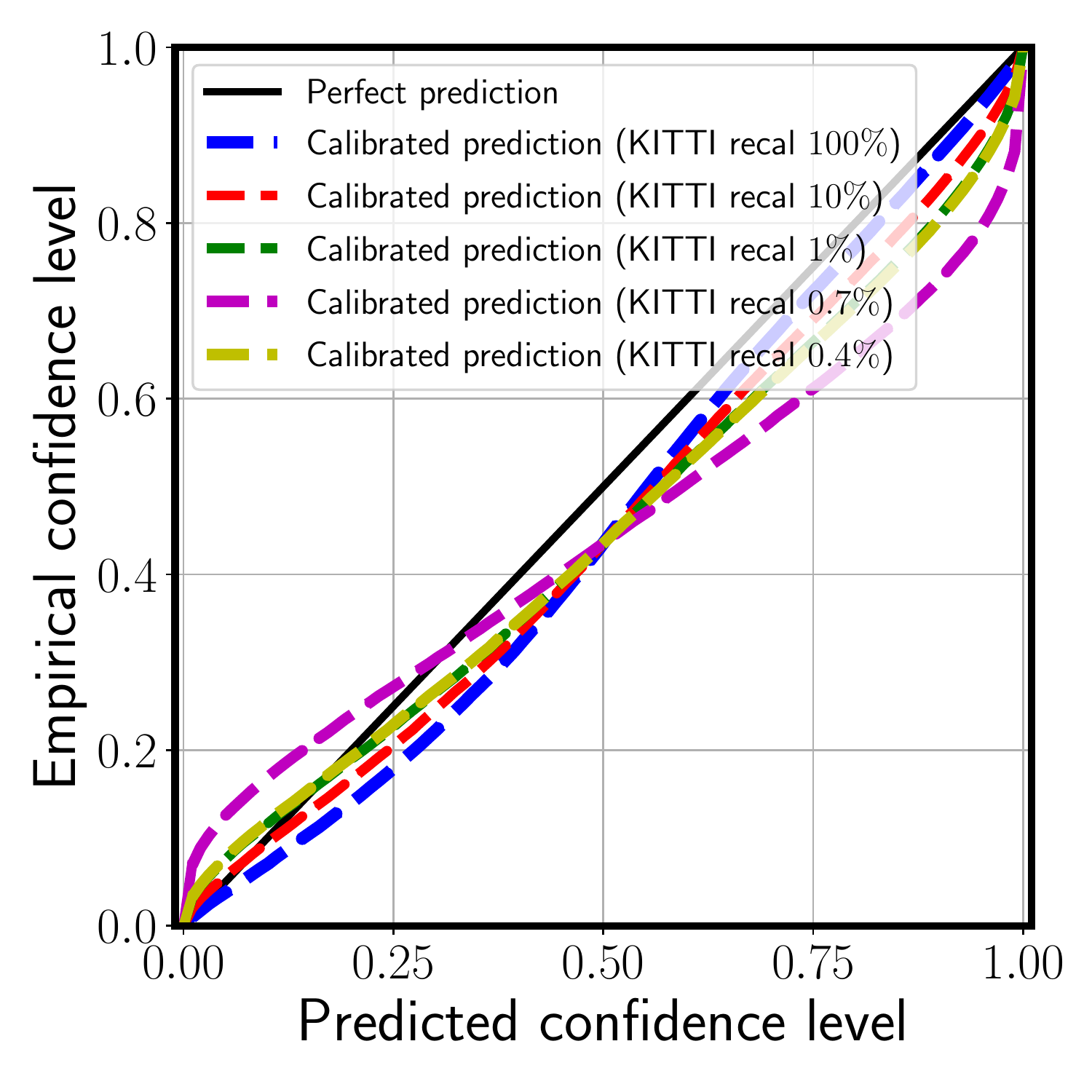}}
		\subfigure[Isotonic Regr.]{\label{fig:num_data_isotopic}\includegraphics[width=0.49\textwidth]{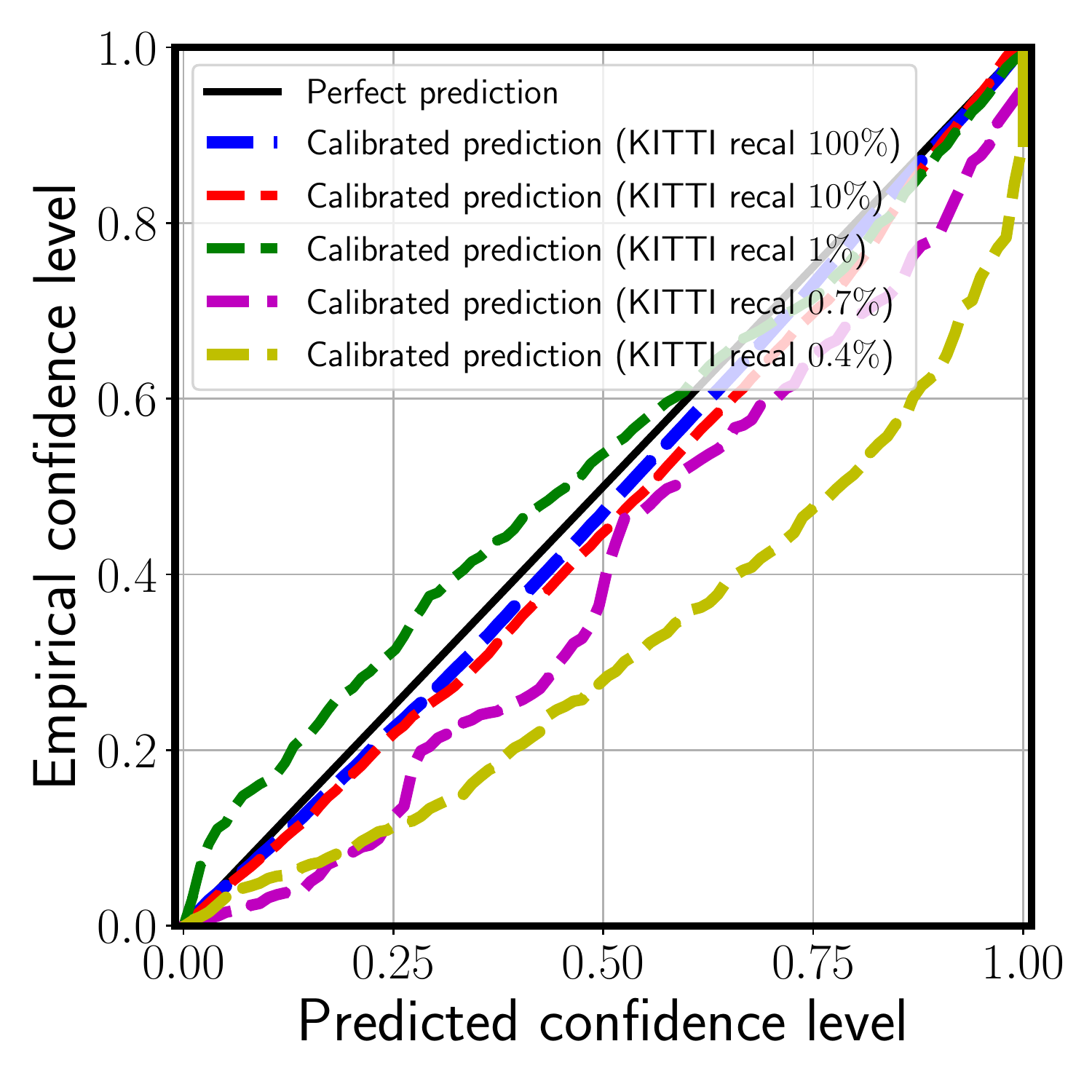}}
		\caption{Evaluating the robustness of the recalibration models against different recalibration dataset sizes. Both methods are optimized with the reduced KITTI \textit{recal} set and evaluated on the KITTI \textit{eval} set.}\label{fig:robustness_num_data}
	\end{minipage}
	\vspace{-1em}
\end{figure}

\subsubsection{Qualitative Observations}
Fig.~\ref{fig:qualitative_observation} illustrates how recalibration models adjust uncertainty estimation. We colorize each detection according to the softmax score, and draw $95\%$ confidence intervals for $dx$ and $dy$ regressions respectively. We use Isotonic Regr. to recalibrate classification uncertainty, and Temp. Scaling for regression uncertainties (in order to keep the Gaussian distribution assumption). From Fig.~\ref{fig:1} and Fig.~\ref{fig:2} we observe that the recalibration model reduces classification scores for the false positive detections (detections 1, 3, 6, 7), and increases classification scores for the true positive detections (detections 2, 4, 5, 8, 9). This is because the uncalibrated network produces over-confident predictions at small softmax scores, and under-confident predictions at bigger softmax scores (Fig.~\ref{fig:calibration_classification}). Fig.~\ref{fig:3} shows that the positional confidence intervals do not fully cover the object ground truths (detections 12, 14, 16), as the uncalibrated network produces over-confident position estimations (Fig.~\ref{fig:calibration_regression}). After recalibration, the confidence intervals are larger such that they fully cover the object positions (Fig.~\ref{fig:4}). Note that obj$15$ is a false negative in ground truth labels.

\subsubsection{Robustness Testing}
We extensively study the robustness of the recalibration methods. Here, we focus on evaluating ``Temp. Scaling" and ``Isotonic Regr.", as their performance highly depends on the recalibration dataset.  First, we optimize both models with the reduced KITTI \textit{recal} set and check their performance on the KITTI \textit{eval} set. Fig.~\ref{fig:robustness_num_data} demonstrates that both methods show good recalibration performance, even with merely $1\%$ of the recalibration data. While ``Isotonic Regr." degrades dramatically with $0.4\%$ of data, ``Temp. Scaling" performs surprisingly well, showing its high robustness against small recalibration dataset size. 

Next, we evaluate the recalibration generalization capability on new test data. In this regard, we train ``Temp. Scaling" and ``Isotonic Regr." with the KITTI \textit{recal} set, and evaluate them on the nuScene dataset~\cite{caesar2019nuscenes}. Tab.~\ref{tab:test_on_nuscene} shows the ECE averaged over all network predictions. The nuScene data significantly differs from the KITTI dataset regarding recording locations, weathers, and sensor setup. Despite that, the recalibration models trained with the KITTI data still halved the averaged ECE in the nuScene data, showing good generalization capability in recalibrating uncertainties. When using only $1\%$ of nuScene data to update the recalibration models, we achieve the best uncertainty recalibration performance.
\begin{table}[htbp]
	\centering
	\resizebox{0.65\linewidth}{!}{
	\begin{tabular}{c | c}
		\rowcolor{lightgray!75}
		\textbf{\texttt{Method}} & \textbf{\texttt{avg. ECE}}  \\ 
		\textbf{\texttt{Uncalibrated (Baseline)}} & $0.218\ \ $  \\
		\textbf{\texttt{Temp. Scaling (KITTI \textit{recal} $100\%$)}}   & $0.126\downarrow$   \\ 
		\textbf{\texttt{Isotonic Regr. (KITTI \textit{recal} $100\%$)}}  & $0.120\downarrow$   \\ 
		\textbf{\texttt{Temp. Scaling (nuScene $1\%$)}}   & $0.078\downarrow$   \\ 
		\textbf{\texttt{Isotonic Regr. (nuScene $1\%$)}}  & $\mathbf{0.030}\downarrow$   \\ 
	\end{tabular}}
	\caption{Averaged Expected calibration errors (ECE) on the nuScene dataset. The recalibration models are optimized using KITTI \textit{recal} set or only $1\%$ of the nuScene data.}
	\label{tab:test_on_nuscene}
	\vspace{-2em}
\end{table}

\section{Conclusion and Discussion}
\label{sec:conclusion}
In this work, we identify that the direct-modeling method, which is a common method to model uncertainty in deep object detectors, produces miscalibrated uncertainties. Based on a probabilistic LiDAR 3D object detector and calibration plots, we study how the course of training affects the uncertainty miscalibration, especially for the bounding box regression task. Then, we propose three practical uncertainty recalibration methods to alleviate such problems. Experiments on both KITTI and nuScene datasets show that our methods estimate well-calibrated uncertainties, are robust against different recalibration dataset sizes and can generalize to new datasets. 

\begin{figure}[tpb]
	\centering
	\begin{minipage}{0.86\linewidth}
		\centering
		\includegraphics[width=1\linewidth]{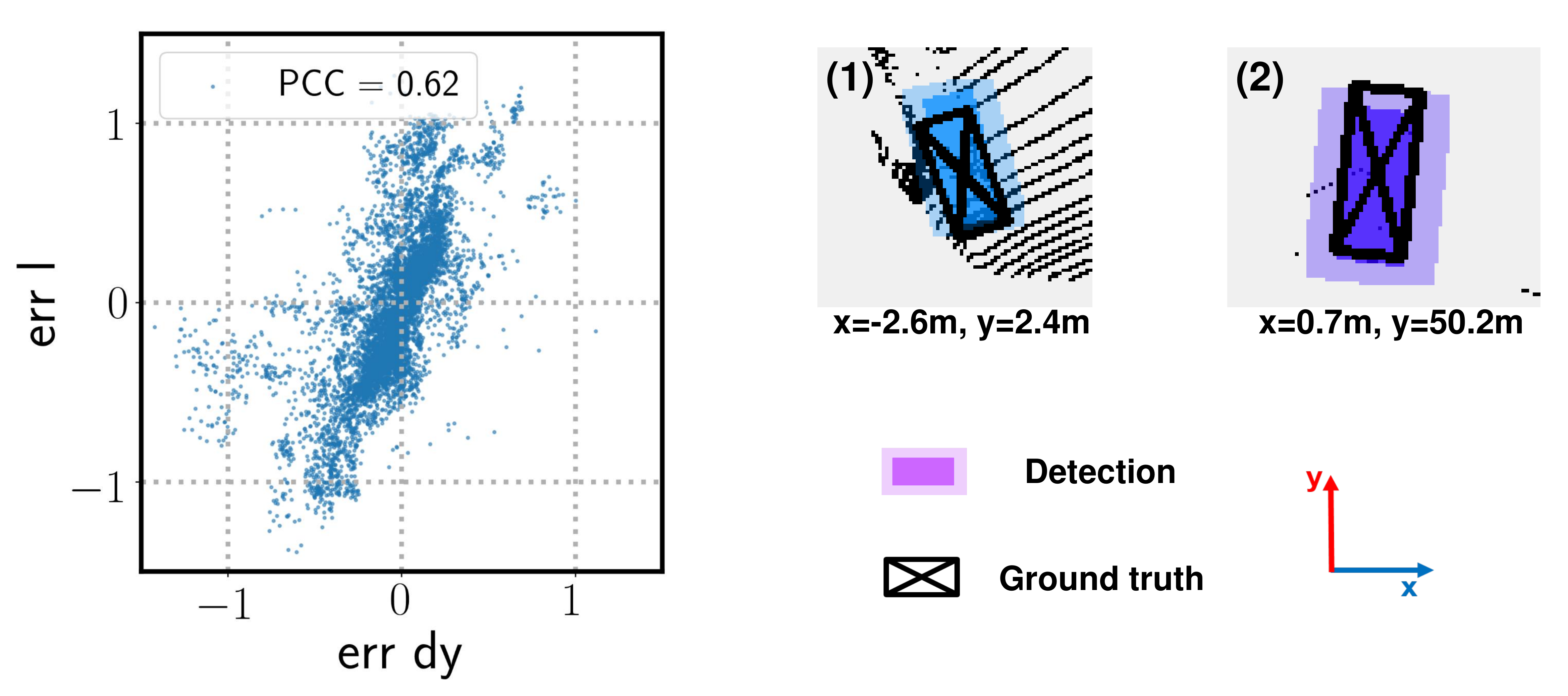}
	\end{minipage}
		\caption{The distribution between the errors of $dy$ (longitudinal position) and $l$ predictions (object length) from the objects which are standing approximately straight in front of the ego-vehicle. They are correlated with $PCC=0.62$, which is shown by two examples.}\label{fig:discussion}
	\vspace{-2em}
\end{figure}

This work calibrates marginal probabilities in the regression task, with the assumption that each regression variable is independently-distributed. However, we find that the regression variables can be dependent in some scenarios as well. Fig.~\ref{fig:discussion} illustrates the distribution between the errors of $dy$ predictions (longitudinal position) and $l$ predictions (object length) from the objects which are standing approximately straight in front of the ego-vehicle, and are facing towards or backwards to the ego-vehicle. The errors of $dy$ and $l$ are highly correlated, with a Pearson Correlation Coefficient of more than $0.6$. We also show two examplary detections, where the object parts which face towards the ego-vehicle are well-localized, but the backsides have large localization errors. In these scenarios, it is necessary to take dependency within regression variables into consideration (instead of estimating and calibrating marginal probabilities), or propose new bounding box encodings that ensures independency. We leave it as an interesting future work. Furthermore, we intend to model uncertainties in multi-modal fusion networks~\cite{feng2019deep} and network quantization~\cite{enderich2019learning}.

\bibliographystyle{IEEEtran}
\bibliography{bibliography}

\end{document}